\documentclass{article}

\usepackage[preprint]{neurips_2026}

\usepackage[utf8]{inputenc} 
\usepackage[T1]{fontenc}    
\usepackage{hyperref}       
\usepackage{url}            
\usepackage{booktabs}       
\usepackage{amsfonts}       
\usepackage{nicefrac}       
\usepackage{microtype}      
\usepackage{xcolor}         

\usepackage{graphicx}
\usepackage{float}
\usepackage{amsmath}

\newif\ifoverleaf
\overleaffalse  

\ifoverleaf
  
\else
  
\fi

\title{Is Inter-Seed Cross-Play Enough? Evaluating the Robustness of Zero-Shot Coordination Algorithms to Implementation Details}

\author{%
  Maksymilian Wolski\thanks{Work done partially while at QuantCo} \\
  ProrokLab\\
  University of Cambridge\\
  United Kingdom \\
  \texttt{} \\
  \And
  Nicholas Hoernle \\
  QuantCo\\
  \texttt{} \\
  \And
  Johannes Forkel \\
  BOLD\footnotemark[2] \\
  University of Oxford\\
  United Kingdom \\
  \texttt{} \\
  \AND
  Jakob Foerster \\
  BOLD\thanks{British Open-Ended Learning and Discovery Lab} \\
  University of Oxford\\
  United Kingdom \\
  \texttt{} \\
}

\begin{document}

\maketitle

\begin{abstract}

      AI agents deployed in real-world settings must be capable of coordinating with humans and other AI agents they have not encountered before.
      Zero-shot coordination (ZSC) algorithms aim to achieve this by specifying high-level learning rules such that independently engineered agents can coordinate with each other at test time.
      Rigorous evaluation of ZSC algorithms remains difficult: ideally, multiple independent implementations of each proposed algorithm must be used, reflecting the variation that arises when independent parties interpret and implement the same specification.
      In practice, however, ZSC algorithms have almost exclusively been evaluated using a single implementation trained across different random seeds, with only a handful of works additionally varying the neural network architecture. This leaves open questions about robustness to specification ambiguities and implementation details.
      In this work, we provide the first systematic evaluation of this robustness. 
      We introduce a new evaluation scheme, cross-implementation cross-play, varying implementation details that prior work has shown to affect the performance of multi-agent reinforcement learning (MARL) algorithms, and we evaluate Other-Play, a popular ZSC algorithm, with this scheme.
      Our findings are encouraging and suggest that, for Other-Play, the standard ZSC evaluation is, in fact, a reasonable proxy for this more thorough cross-implementation evaluation.

\end{abstract}

\section{Introduction}

    As AI agents become more capable, they will increasingly be deployed in settings where they must cooperate with humans and with other agents whose implementations, training procedures, and design choices may be unknown.
    Ensuring that agents can achieve such cooperation with unfamiliar partners is a central challenge in cooperative multi-agent reinforcement learning (MARL): off-the-shelf MARL algorithms typically lead to agents developing arbitrary conventions which are effective within the training population but lead to miscoordination with unfamiliar partners at test time \citep{hu_other-play_2020, treutlein_new_2021, hu_off-belief_2021, dizdarevic_ad-hoc_2025}.

    Zero-shot coordination (ZSC) \citep{hu_other-play_2020, treutlein_new_2021} attempts to address this problem by designing algorithms that, when implemented independently lead to agents that can coordinate successfully with each other, even without prior interaction during training.
    The evaluation of ZSC algorithms remains challenging.
    Ideally, a proposed algorithm is assessed by having multiple independent parties implement it from scratch, training agents from each implementation, and measuring cross-play performance between these agents.
    This would capture the variation introduced by ambiguities in the algorithmic specification and by low-level implementation choices, but is prohibitively expensive in practice.
    Consequently, in the ZSC literature \citep{cui_k-level_2021, hu_off-belief_2021, lupuTrajectoryDiversityZeroShot2021, cui_off-team_2022, muglich_equivariant_2022, muglich_expected_2025}, the standard practice has been to train a single implementation across multiple random seeds, treating inter-seed cross-play as a proxy for cross-implementation performance.

    We provide an empirical evaluation of this simplification.
    Varying the PPO \citep{schulman_proximal_2017} implementation details discussed in \citet{huang_37_2022}, we construct multiple implementations of the same ZSC algorithm by varying code-level details, simulating the variation expected across independent implementations. Our results show no meaningful gap between inter-seed and cross-implementation cross-play performance, providing evidence that the standard evaluation practice is a reasonable simplification.

\section{Background}

    \subsection{Cooperative Multi-Agent Reinforcement Learning}
  We consider a setting with multiple agents acting in a partially-observable environment and optimizing for the same objective, formalized as a Decentralized Partially Observable Markov Decision Process (Dec-POMDP) \citep{oliehoek_dec-pomdps_2007}. A Dec-POMDP is a tuple $(N, \mathcal{S}, \{\mathcal{A}^i\}_{i=1}^{N}, \mathcal{P}, \{\mathcal{O}^i\}_{i=1}^{N}, \Omega, \mu_0, R, \gamma)$, where $N$ is the number of agents, $\mathcal{S}$ is the state space, $\mathcal{A}^i$ and $\mathcal{O}^i$ are the action and observation spaces for agent $i$, $\mathcal{P}$ is the transition function, $\Omega$ is the observation function, $\mu_0$ is the initial state distribution, $R$ is the shared reward function, and $\gamma \in [0,1)$ is the discount factor.
  
 Rather than observing the full state, each agent receives only a local observation $o^i_t$ and maintains a local history $h^i_t = (o^i_0, a^i_0, \dots, o^i_{t-1}, a^i_{t-1}, o^i_t)$, where each agent's policy $\pi^i$ maps local histories to a distribution over actions, with actions sampled as $a_t^i \sim \pi^i(\cdot \mid h^i_t)$. The reward function $R$ is shared across all agents, and all agents jointly aim to maximize the expected discounted return
$J(\boldsymbol{\pi}) := \mathbb{E}_{\boldsymbol{\pi}} \left[ \sum_{t=0}^{\infty} \gamma^t R(s_t, \boldsymbol{a}_t) \right],$
where we use boldface $\boldsymbol{\pi} = \left(\pi^1,\dots,\pi^N\right)$ and $\boldsymbol{a} = \left(a^1,\dots,a^N\right)$ to denote the joint policy and joint actions, respectively, distinguishing them from their single-agent counterparts.

An optimal policy for a Dec-POMDP is one that maximizes the joint expected return,
$\boldsymbol{\pi}^* \in \left\{ \boldsymbol{\pi}' : J(\boldsymbol{\pi}') = \max_{\boldsymbol{\pi}} J(\boldsymbol{\pi}) \right\}.$
In the setting of self-play, agents are both trained and evaluated together. We refer to self-play score as
$\operatorname{SP}(\boldsymbol{\pi}) = J(\boldsymbol{\pi})$.
        
    \subsection{Zero-Shot Coordination}
    
        ZSC, as introduced in \citet{hu_other-play_2020} and further formalized by \citet{treutlein_new_2021}, aims to overcome the problem of agents over-optimizing for their training partners. It seeks high-level, implementation-agnostic algorithms under which independently trained agent populations coordinate successfully with each other at test time, without any prior communication between parties beyond agreement on the algorithm. An example of such an algorithm is Other-Play \citep{hu_other-play_2020}, which modifies the training objective to prevent agents from breaking environment symmetries, promoting more robust coordination.
    
    \subsection{Evaluation of Zero-Shot Coordination}

        To evaluate a ZSC algorithm $\mathcal{L}$, previous papers, e.g. \citet{cui_k-level_2021, hu_off-belief_2021, lupuTrajectoryDiversityZeroShot2021, cui_off-team_2022, muglich_equivariant_2022, muglich_expected_2025} typically trained a single implementation $\mathcal{L}_k$ across $S$ random seeds, producing policies $\boldsymbol{\pi}_{k_1}, \boldsymbol{\pi}_{k_2}, \ldots, \boldsymbol{\pi}_{k_S}$. For a two-player game, the cross-play score between two joint policies $\boldsymbol{\pi}_1, \boldsymbol{\pi}_2$ is defined as:
        \begin{equation}
        \operatorname{XP}(\boldsymbol{\pi}_1, \boldsymbol{\pi}_2) = \frac{1}{2} \left( J((\pi^1_1, \pi^2_2)) + J((\pi^1_2, \pi^2_1)) \right).
        \end{equation}

    Given $S$ policies $\{\boldsymbol{\pi}_{k_i}\}_{i=1}^{S}$ trained from implementation $\mathcal{L}_k$, we construct the $S \times S$ cross-play matrix whose $(i,j)$-th entry is $J((\pi^1_{k_i}, \pi^2_{k_j}))$. The average self-play and cross-play scores are then the means of its diagonal and off-diagonal entries, respectively:
        \begin{align}
            \overline{\operatorname{SP}}(\mathcal{L}_k) &= \frac{1}{S} \sum_{i=1}^{S} \operatorname{XP}(\boldsymbol{\pi}_{k_i}, \boldsymbol{\pi}_{k_i}) \\
            \label{eq:xp}
            \overline{\operatorname{XP}}(\mathcal{L}_k) &= \frac{1}{S(S-1)} \sum_{i \neq j} \operatorname{XP}(\boldsymbol{\pi}_{k_i}, \boldsymbol{\pi}_{k_j})
        \end{align}
        Note: $\boldsymbol{\pi}_{k_i}$, $i \in [S]$, are trained from \emph{the single implementation $\mathcal{L}_k$}. 
        ZSC aims to find algorithms that maximize $\overline{\operatorname{XP}}$.

        The difference $\overline{\operatorname{SP}}-\overline{\operatorname{XP}}$, between average self-play and average cross-play (self-play-cross-play gap), is used to determine if the trained policies are zero-shot compatible.
        A small gap indicates agents coordinate as well with unseen partners as with seen partners, while a large gap reveals over-reliance on arbitrary conventions formed during training.

\begin{figure*}
    \centering
    \includegraphics[width=\textwidth]{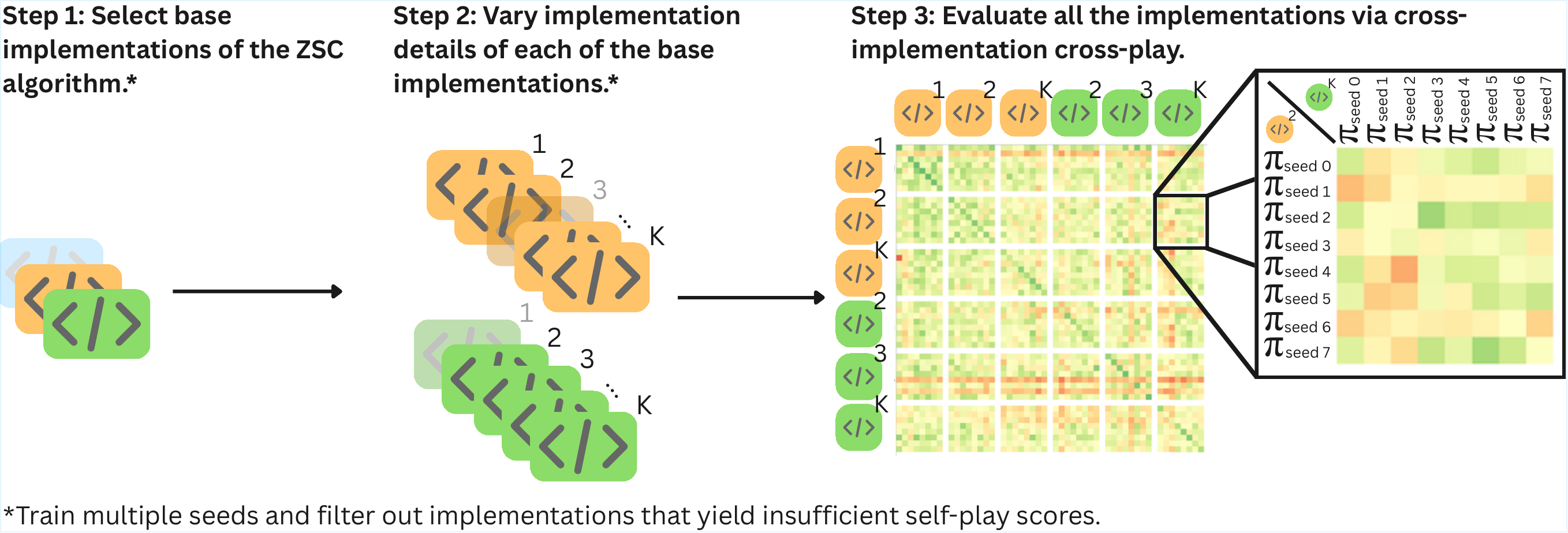}
    \caption{Overview of the cross-implementation cross-play (XIXP) evaluation procedure. 
For a given ZSC algorithm and environment, we first generate base implementations 
that achieve sufficient self-play performance, then apply a range of implementation 
details to each. The resulting implementations are finally evaluated via cross-implementation 
cross-play.}
    \label{fig:hero}
\end{figure*}

\section{Related Work}

Previous ZSC algorithms have been evaluated by only accounting for the architecture of the neural network used \citep{hu_other-play_2020} or without considering implementation details at all \citep{cui_k-level_2021, hu_off-belief_2021, lupuTrajectoryDiversityZeroShot2021, cui_off-team_2022, muglich_equivariant_2022, muglich_expected_2025}.
A recent work that goes beyond network architecture is \citet{forkel_high_2025}, which showed that the entropy regularization coefficient of independent PPO (IPPO) \citep{witt_is_2020}, as well as the $\lambda_\text{GAE}$ coefficient \citep{schulman_high-dimensional_2018} can meaningfully affect the cross-play performance of IPPO-trained agents.
Our work builds on this finding by considering multiple implementation details simultaneously and examines whether ZSC algorithms remain robust under such variation.

Conceptually close to ours is the work of \citet{lucas_any-play_2022}, who evaluate ZSC algorithms in the inter-algorithm cross-play setting, where policies generated by different ZSC algorithms are paired together. 
They propose ``Any-play'' as an improved ZSC algorithm designed for inter-algorithm cross-play. 
In contrast, our work assumes the formalization of ZSC proposed by \citet{treutlein_new_2021}, in which the different parties agree on the ZSC algorithm used for generating agents, sidestepping the inter-algorithm cross-play problem. 
Our focus is on ``cross-implementation cross-play'', which varies the implementation of the ZSC algorithm rather than the algorithm chosen for evaluation.

The implementation details we vary are grounded in prior work on PPO \citep{schulman_proximal_2017}. Most directly relevant is \citet{huang_37_2022}, which studies multiple implementations of this single-agent reinforcement learning algorithm and lists the details that matter for its performance across multiple environments. 
\citet{yu_surprising_2022} similarly highlights important design decisions when adapting PPO to the multi-agent setting. 
We draw on both works to select the implementation details used in this study.

\section{Methodology}

We introduce \emph{cross-implementation cross-play} (XIXP) which evaluates ZSC algorithms across $K$ implementations $\{\mathcal{L}_k\}_{k=1}^{K}$, obtained by varying implementation details of the base MARL algorithm. 
XIXP can be seen as an extension of standard cross-play evaluation in which policies differ not only in random seed but also in the underlying implementation of the ZSC-augmented algorithm.

\subsection{Definition of Cross-Implementation Cross-Play}

For two implementations $\mathcal{L}_k$ and $\mathcal{L}_m$, each trained across $S$ random 
seeds, we define cross-implementation cross-play (XIXP) as the average cross-play score over all pairs of 
policies drawn from each implementation:\begin{equation} \label{eq:xxp_km}
    \begin{aligned}
        \operatorname{XIXP}(\mathcal{L}_k, \mathcal{L}_m) &:= \frac{1}{S^2}\sum_{i,j \in [S]} 
        \operatorname{XP}(\boldsymbol{\pi}_{k_i}, \boldsymbol{\pi}_{m_j}), \quad k \neq m,\\
        \operatorname{XIXP}(\mathcal{L}_k, \mathcal{L}_k) &:= \overline{\operatorname{XP}}(\mathcal{L}_k).
    \end{aligned}
\end{equation}
We summarize performance across all implementation pairs via the average XIXP score and within-implementation cross-play (WIXP) score:
\begin{equation}
    \begin{aligned}
        \overline{\operatorname{XIXP}}(\mathcal{L}) &:= \frac{1}{K(K-1)} \sum_{\substack{k,m \in [K] \\ k \neq m}} \operatorname{XIXP}(\mathcal{L}_k, \mathcal{L}_m),\\
        \overline{\operatorname{WIXP}}(\mathcal{L}) &:= \frac{1}{K} \sum_{k \in [K]} \operatorname{XIXP}(\mathcal{L}_k, \mathcal{L}_k).
    \end{aligned}
\end{equation}

Analogous to the self-play-cross-play gap,
$\overline{\operatorname{WIXP}} - \overline{\operatorname{XIXP}}$, indicates whether 
cross-implementation miscoordination exists beyond what is already captured by inter-seed XP. 

\subsection{Experimentation Details}
\label{sec:exp_details}

We evaluate the robustness of Other-Play \citep{hu_other-play_2020} in the Yokai Learning Environment \citep{ruhdorfer_yokai_2026-1}, a new benchmark to test ZSC, recently proposed as an alternative to Hanabi \citep{bardHanabiChallengeNew2020a}. As the base MARL algorithm, we use IPPO, which achieves strong self-play performance in Yokai and allows us to directly apply the implementation details described in \citet{huang_37_2022}.

As shown in Figure~\ref{fig:hero}, we first evaluate Other-Play with IPPO on three neural network architectures (GRU, LSTM, and feedforward) with two entropy coefficients (0.01 and 0.05). GRU and LSTM substantially outperform feedforward networks, and the higher entropy coefficient consistently improves performance across all architectures.\footnote{The self-play results are summarized in Appendix Figure~\ref{fig:arch_comparison-sp_xp}}

Step 2 is to generate further implementations by varying implementation details. We varied: $\lambda_{\text{GAE}}$, the learning rate schedule (annealing vs.\ constant), gradient norm clipping, value function clipping, weight initialization (Xavier vs.\ orthogonal), number of hidden layers, number of minibatches, and the discount factor $\gamma$.

To ensure that XIXP scores reflect meaningful coordination rather than the failure of incompetent policies, we discard any implementation whose self-play score falls below a threshold, retaining only those that produce competent policies. 
For estimation of the standard error of the various average XP scores, we use the estimators proposed by \citet{forkel_high_2025} in which each seed contributes exactly one 
inter-seed pairing per implementation. This preserves the independence assumption required for valid confidence intervals, which is violated by the common practice of evaluating each seed against all others, as is done in Equation~\ref{eq:xp}. 

We further observe that XP scores tend to be multimodal rather than uniformly 
distributed, necessitating a sufficient number of seeds to reliably capture all modes.
As training a single seed of IPPO in Yokai takes up to 9 hours on an NVIDIA A40 GPU, we train 8 seeds per implementation.

\section{Results}

In total, we trained 176 policies across 22 implementations. Eleven of the implementations yielded policies with insufficient self-play performance and were discarded ($\overline{\operatorname{SP}}(\mathcal{L}_k)~<~5$). The
remaining 11 implementations, based on IPPO with GRU and LSTM neural network architectures, form the basis of our XIXP
analysis.\footnote{The full self-play and cross-play scores are shown in Figure~\ref{fig:all_gru_and_lstm_op-sp_xp} in the Appendix}

\begin{figure}
    \centering
    \includegraphics[width=0.6\linewidth]{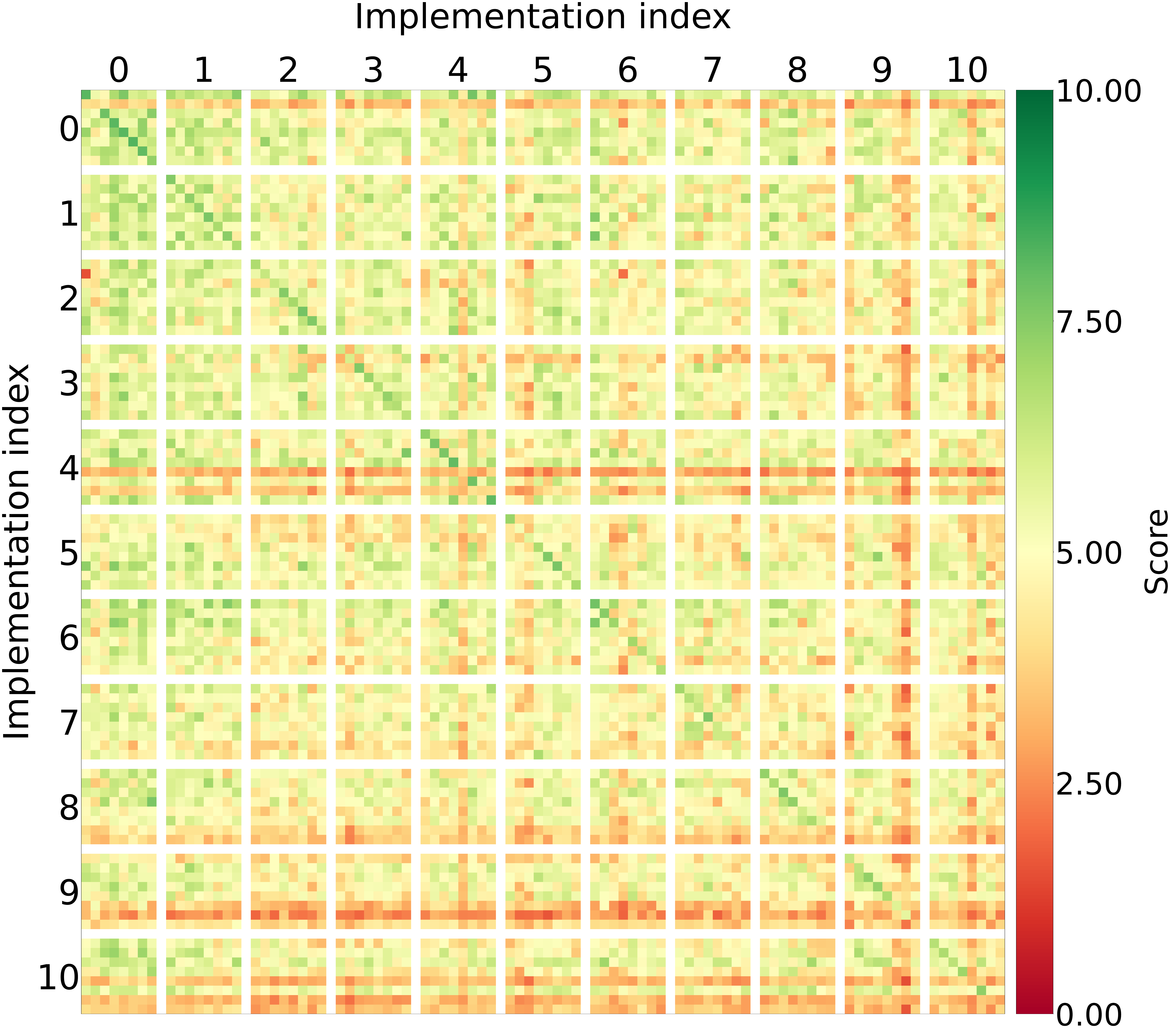}
    \caption{XIXP matrix of Other-Play trained with IPPO across GRU and LSTM architectures and various implementation details, restricted to those achieving sufficient self-play performance. Each implementation was trained with 8 seeds. Implementation indices are used for readability; full implementation names are shown in Figure~\ref{fig:xxp_plot-xxp_heatmap_full} in the Appendix.
    Scores across implementations are comparable to scores within-implementations, indicating no meaningful cross-implementation gap.}
    \label{fig:xxp_plot-xxp_heatmap_indexed}
\end{figure}

The $11 \times 11$ XIXP matrix in Figure~\ref{fig:xxp_plot-xxp_heatmap_indexed} presents cross-play scores for all implementation pairs considered in this study. The diagonal tiles represent self-play scores for each policy; the off-diagonal tiles represent cross-play between seeds, grouped by implementation pair. 
Visually, we observe no meaningful difference between within-implementation cross-play and across-implementation cross-play, indicating that policies coordinate 
equally well whether paired within the same implementation or across different ones. 

The aggregate scores in Table~\ref{tab:gru-sxp_xxp} confirm this observation quantitatively. We observe no $\overline{\operatorname{WIXP}} - \overline{\operatorname{XIXP}}$ gap, suggesting that the implementation details we varied do not 
meaningfully affect cross-play performance. Thus, these results provide empirical support for the standard 
ZSC evaluation practice of using inter-seed cross-play as a proxy for cross-implementation 
coordination.

\begin{table}[ht]
  \centering
  \begin{tabular}{lrrr}
    \toprule
    & Mean & 95\,\% CI low & 95\,\% CI high \\
    \midrule
    $\overline{\operatorname{WIXP}}$ & 4.8892 & 4.6109 & 5.1675 \\
    $\overline{\operatorname{XIXP}}$ & 4.8487 & 4.7548 & 4.9427 \\
    \bottomrule
  \end{tabular}
  \caption{Aggregate $\overline{\operatorname{WIXP}}$ and $\overline{\operatorname{XIXP}}$ scores for the selected implementations presented in Figure~\ref{fig:xxp_plot-xxp_heatmap_indexed}. The overlapping confidence intervals and near-identical means confirm no statistically significant $\overline{\operatorname{WIXP}} - \overline{\operatorname{XIXP}}$ gap.}
    \label{tab:gru-sxp_xxp} 
\end{table}
Figure~\ref{fig:xxp_plot-xxp_heatmap_indexed} also illustrates a subtler point about the importance of training a sufficient number of seeds 
and adopting our evaluation scheme discussed in Section~\ref{sec:exp_details}. In the GRU+OP (no VF clip) implementation (index 9 in Figure~\ref{fig:xxp_plot-xxp_heatmap_indexed}), 3 seeds 
produced policies that were noticeably weaker than the rest. With fewer seeds, these 
could distort aggregate metrics. Moreover, under the conventional evaluation scheme, where 
each seed plays against all others, these 3 seeds would participate in a disproportionate 
number of pairings, potentially leading to a misleading conclusion that the implementation 
itself yields poor cross-play performance.

\section{Conclusion}

We introduced cross-implementation cross-play (XIXP), an evaluation framework for measuring the robustness of 
ZSC algorithms to implementation details. By training multiple implementations of 
Other-Play, obtained by varying code-level details of the underlying IPPO algorithm, we 
simulated the diversity one would expect from the implementations generated by independent parties. Our results show 
no meaningful $\overline{\operatorname{WIXP}} - \overline{\operatorname{XIXP}}$ gap, suggesting that inter-seed cross-play is a reliable proxy for 
cross-implementation evaluation and providing empirical support for the standard evaluation 
practice in ZSC research.

The computational cost of XIXP evaluation 
restricted our analysis to a single environment (Yokai), a single base algorithm (IPPO), and 
a single ZSC algorithm (Other-Play). Whether these findings generalize to other 
environments, algorithms such as Q-learning, and ZSC algorithms such as Off-Belief Learning remains an open question. Extending this study along these dimensions is a natural direction for future work.

\newpage
\section*{Acknowledgments}
This work was supported by the Engineering and Physical Sciences Research Council (grant number UKRI5149). This project also received support from QuantCo, in the form of compute resources and funding during an internship. J. Foerster is partially funded and J. Forkel is fully funded by the UKRI grant EP/Y028481/1
(originally selected for funding by the ERC). We thank Darius Muglich and Constantin Ruhdorfer for helpful discussions.

\bibliographystyle{unsrtnat}
\bibliography{references}

\appendix
    
\twocolumn

\section{Supplementary Figures}
\begin{figure}[H]
    \centering
    \includegraphics[width=1\linewidth]{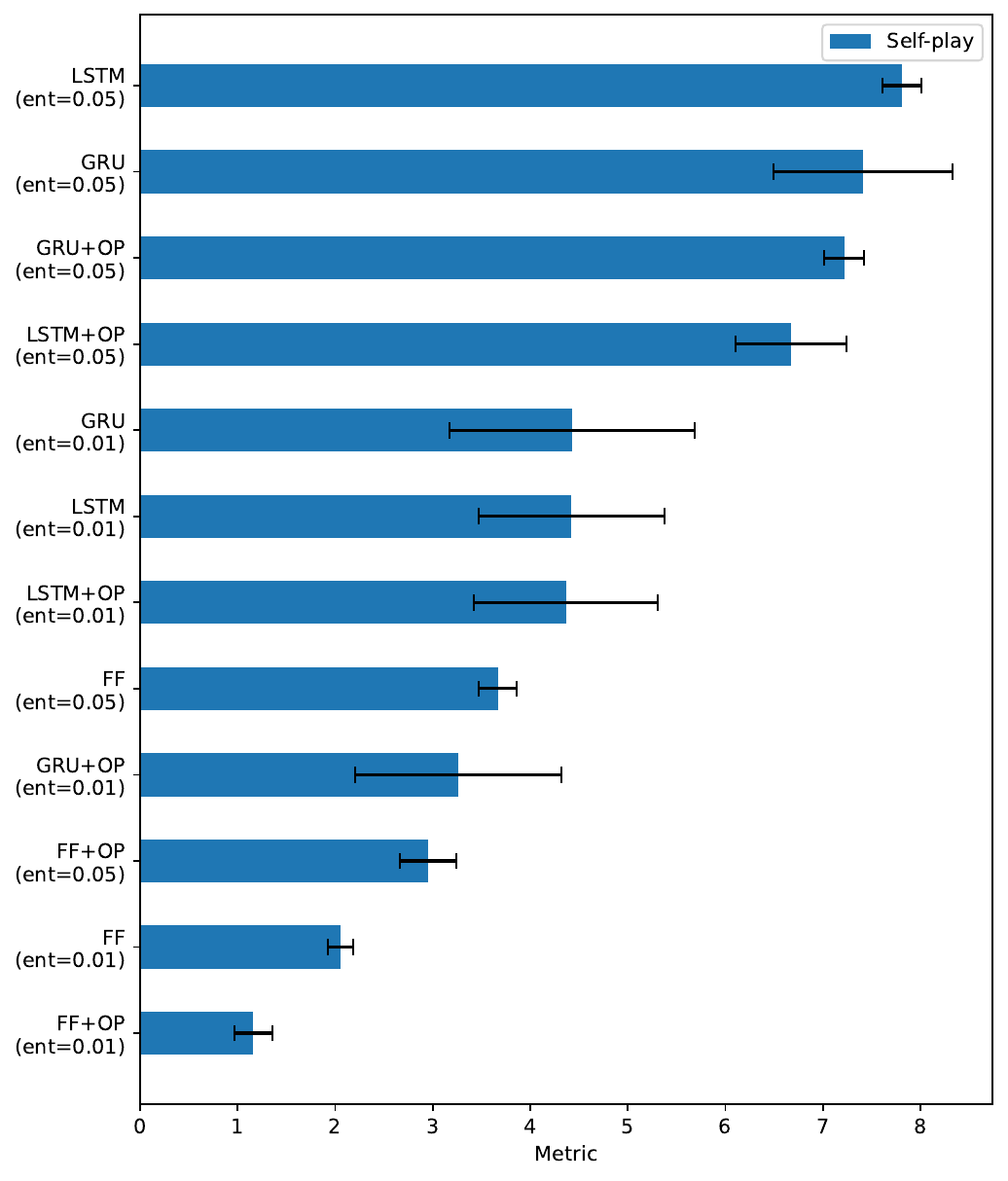}
    \caption{Self-play scores of IPPO in Yokai across three neural network architectures 
(GRU, LSTM, feedforward), with and without Other-Play, at entropy coefficients 0.01 
and 0.05. A higher entropy coefficient (0.05) consistently improves self-play 
performance across all architectures. Only IPPO with the LSTM and GRU architectures generate competent policies.}
    \label{fig:arch_comparison-sp_xp}
\end{figure}

\begin{figure}
    \centering
    \includegraphics[width=1\linewidth]{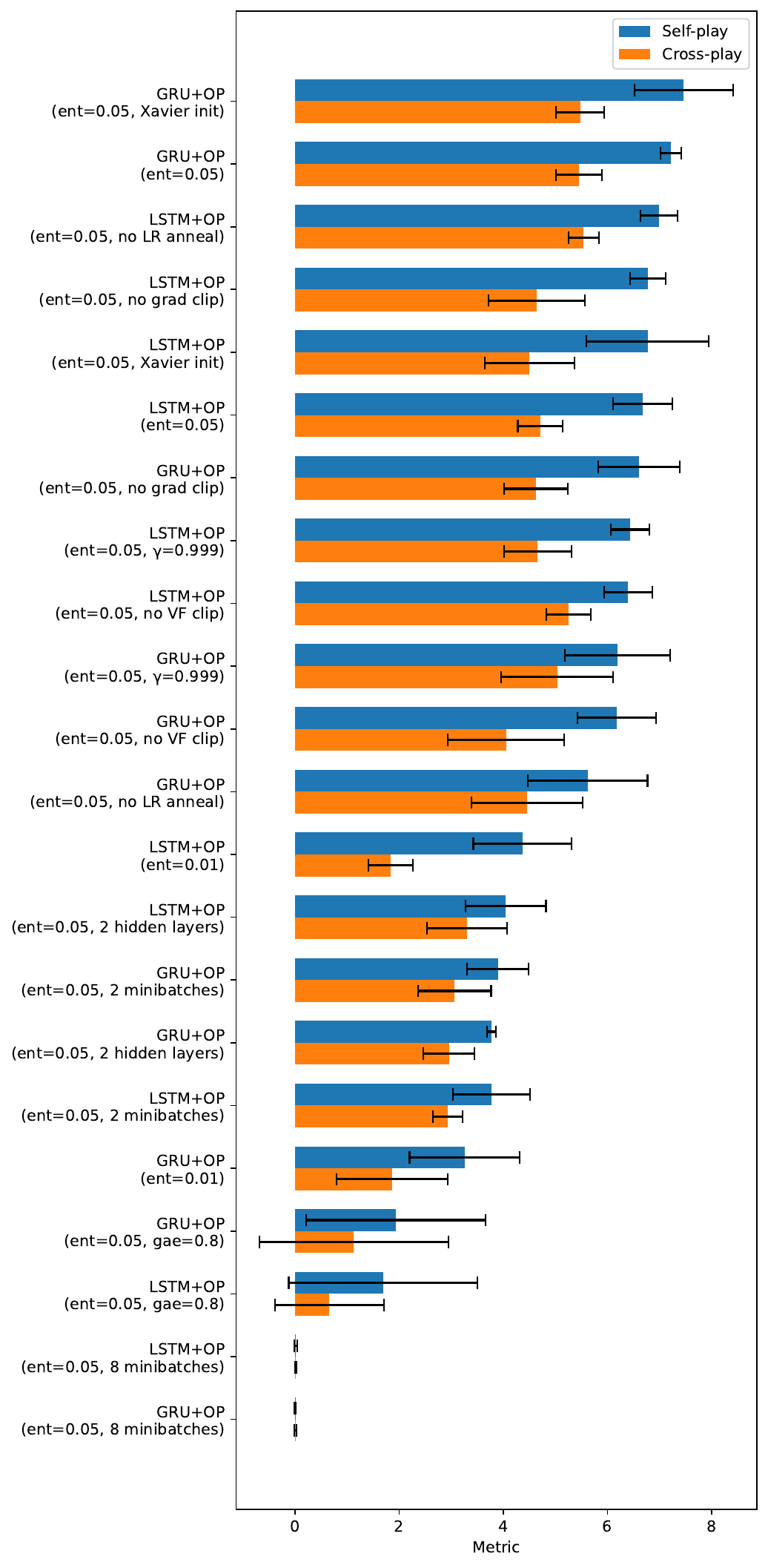}

    \caption{Self-play and cross-play scores for all IPPO implementations using GRU and 
LSTM architectures trained with Other-Play, across all implementation details 
considered in this study (prior to filtering). Several implementation details produce 
policies with insufficient self-play performance and are excluded from the XIXP 
analysis.}
    \label{fig:all_gru_and_lstm_op-sp_xp}
\end{figure}
\onecolumn
\begin{figure}
    \centering
    \includegraphics[width=1\linewidth]{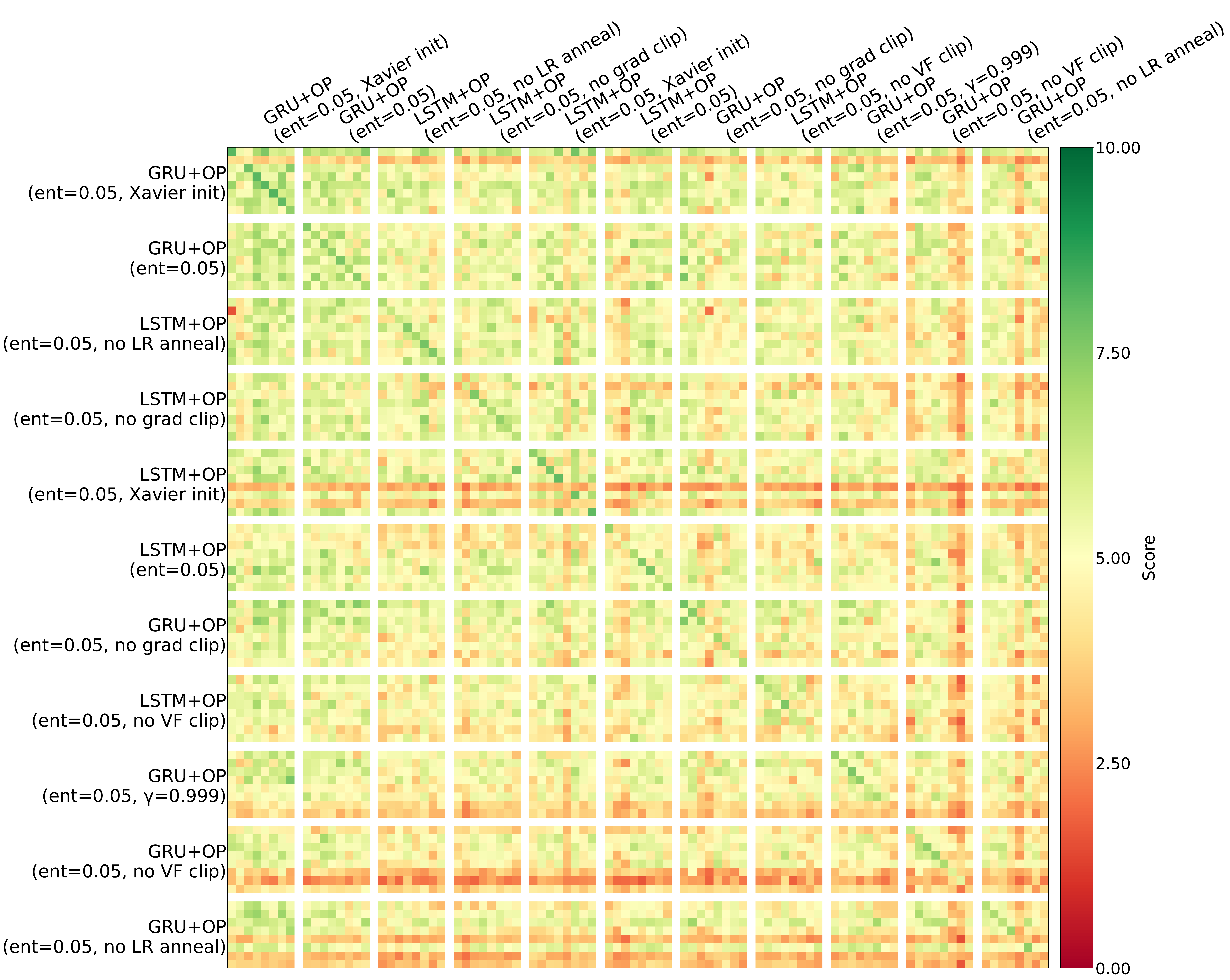}

    \caption{XIXP matrix of Other-Play trained with IPPO across GRU and LSTM architectures and various implementation details, restricted to those achieving sufficient self-play performance. Each implementation was trained with 8 seeds. Implementation names are shown in full; the indexed 
version appears as Figure~\ref{fig:xxp_plot-xxp_heatmap_indexed} in the main text.}
    \label{fig:xxp_plot-xxp_heatmap_full}
\end{figure}

\end{document}